\RequirePackage{fix-cm}
\documentclass[twocolumn]{svjour3}          
\smartqed  

\usepackage{graphicx}
\usepackage{amsmath,amssymb} 
\usepackage{color}
\usepackage{multirow}
\usepackage{array}
\usepackage{xspace}
\usepackage{pifont}
\usepackage{subfig}

\def\eg{\textit{e.g.}\xspace}
\def\ie{\textit{i.e.}\xspace}

\newcommand{\set}[1]{\mathcal {#1}}
\newcommand{\con}[1]{\mathsf {#1}}
\newcommand{\mat}[1]{\mathtt {#1}}

\newcommand{\tochange}[1]{{#1}}

\DeclareMathOperator\sign{sgn}

\newcommand{\myparagraph}[1]{
\vspace{10pt}
\noindent\textbf{#1}
}

\makeatletter
\newlength\eqcol@newlen
\newlength\eqcol@oldlen
\let\eqcol@bc\hfil
\let\eqcol@ec\hfil
\let\eqcol@br\hfil
\let\eqcol@el\hfil
\newcolumntype{e}[1]{%
  >{\setbox0\hbox\bgroup}#1%
  <{\egroup
    \ifdim\wd0<\eqcol@newlen\else\global\eqcol@newlen\wd0\fi
    \ifdim\wd0<\eqcol@oldlen\else\global\eqcol@oldlen\wd0\fi
    \hbox to \eqcol@oldlen{%
      \csname eqcol@b#1\endcsname
      \box0 %
      \csname eqcol@e#1\endcsname
    }%
  }%
}
\newcount\eqcol@count
\def\eqcolRead{%
  \global\advance\eqcol@count1 %
  \eqcol@oldlen5em\relax
  \csname eqcol@def@\romannumeral\eqcol@count\endcsname
}
\def\eqcolWrite{%
  \immediate\write\@auxout{%
  \gdef\expandafter\noexpand\csname eqcol@def@\romannumeral\eqcol@count\endcsname
    {\global\eqcol@oldlen\the\eqcol@newlen\relax}%
  }%
  \global\eqcol@newlen0pt\relax
}
\let\eqcol@old@tabular\tabular
\def\tabular{\eqcolRead\eqcol@old@tabular}
\let\eqcol@old@endtabular\endtabular
\def\endtabular{\eqcol@old@endtabular\eqcolWrite}
\makeatother

\begin{document}

\title{{Boosting Binary Masks for Multi-Domain Learning through Affine Transformations}
}


\author{Massimiliano Mancini \and Elisa Ricci \and Barbara Caputo \\\and Samuel Rota Bul\'o
}


\institute{M. Mancini \at  Sapienza University of Rome, Rome, Italy              \\\email{mancini@diag.uniroma1.it}           
           \and
           E. Ricci \at Fondazione Bruno Kessler, Trento, Italy
           \at University of Trento, Trento, Italy \\\email{eliricci@fbk.eu}
            \and
             B. Caputo \at Politecnico di Torino, Turin, Italy
             \at Italian Institute of Technology, Turin, Italy 
\\\email{barbara.caputo@polito.it}
\and  
S. Rota Bul\'o \at Mapillary Research, Graz, Austria\\\email{samuel@mapillary.com}
}


\maketitle

\begin{abstract}
In this work, we present a new, algorithm for multi-domain learning. Given a pretrained architecture and a set of visual domains received sequentially, the goal of multi-domain learning is to produce a single model performing a task in all the domains together. 
Recent works showed how we can address this problem by masking the internal weights of a given original conv-net through learned binary variables. 
In this work, we provide a general formulation of binary mask based models for multi-domain learning by affine transformations of the original network parameters. Our formulation obtains significantly higher levels of adaptation to new domains, achieving performances comparable to domain-specific models while requiring slightly more than 1 bit per network parameter per additional domain. Experiments on two popular benchmarks showcase the power of our approach, achieving performances close to state-of-the-art methods on the Visual Decathlon Challenge. 
 
\end{abstract}

\section{Introduction}
\label{intro}


{A crucial requirement for visual systems is the ability to adapt an initial pretrained model to novel application domains. 
Achieving this goal requires facing multiple challenges. First, 
{learning a new domain should not negatively affect} the performance on old domains. {Secondly, we should avoid adding many parameters to the model for each new domain that we want to learn, to ensure scalability.} 
In this context, while deep learning algorithms have achieved impressive results on many computer vision benchmarks \cite{krizhevsky2012imagenet,he2016deep,girshick2014rich,long2015fully}, mainstream approaches for adapting deep models to novel domains tend to suffer from the problems  mentioned above.} In fact, fine-tuning a given architecture to new data does produce a powerful model on the novel domain, at the expense of degraded performance on the old ones, resulting in the well-known phenomenon  of  catastrophic forgetting \cite{french1999catastrophic,goodfellow2013empirical}. At the same time, replicating the network parameters and training a separate network for each domain is a powerful approach that preserves performances on old domains, but at the cost of an explosion of the network parameters \cite{rebuffi2017learning}.

{Different works addressed these problems by either considering losses encouraging the preservation of the current weights \cite{li2017learning,kirkpatrick2017overcoming} 
or by designing domain-specific network parameters \cite{rusu2016progressive,rebuffi2017learning,rosenfeld2017incremental,mallya2017packnet,mallya2018piggyback}. Interestingly, in \cite{mallya2018piggyback,mancini2018adding} the authors showed that an effective strategy for achieving good multi-domain learning performances with a minimal increase in terms of network size
is to create a binary mask for each domain. In \cite{mallya2018piggyback} this mask is then multiplied by the main network weights, determining which of them are useful for addressing the new domain. Similarly, in \cite{mancini2018adding} the masks are used as a scaled additive component to the network weights.} 

\begin{figure*}[t]
\centering
 \includegraphics[width=0.7
 \textwidth]{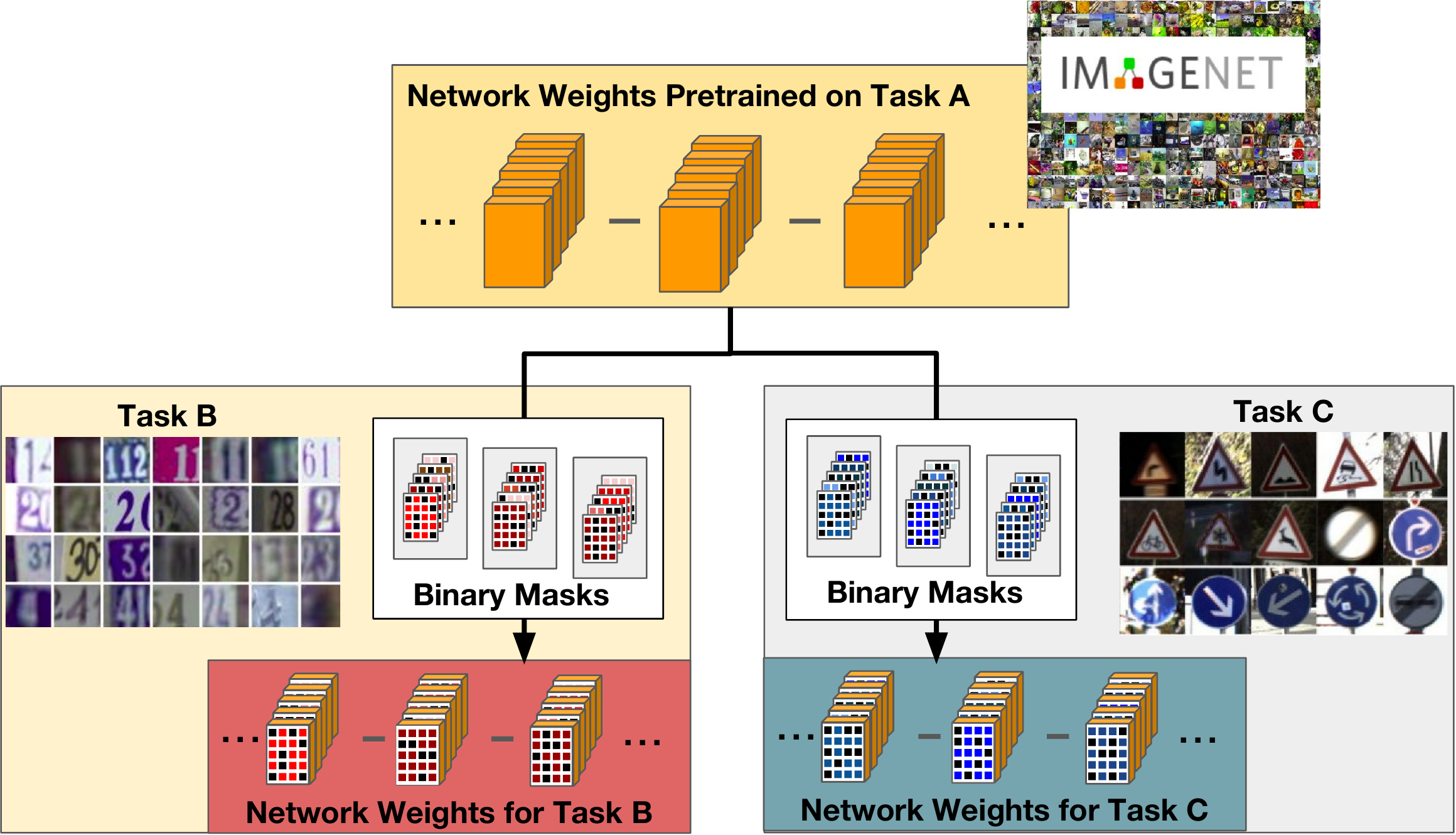} 
  \caption{Idea behind our approach. A network pretrained on a given recognition domain A (\ie ImageNet) can be extended to tackle other recognition domains B (\eg digits) and C (\eg traffic sign)  by simply transforming the network weights (orange cubes) through domain-specific binary masks (colored grids).}
  \label{fig:teaser-masks}
  \end{figure*}

{In this work, we take inspiration from \cite{mallya2018piggyback,mancini2018adding}, formulating multi-domain learning as the problem of learning a {transformation} of a \emph{baseline}}, pretrained network, in a way to maximize the performance on a new domain. Importantly, the {transformation} should be compact in the sense of limiting the number of additional parameters required with respect to the baseline network.
To this extent, we apply an affine transformation to each convolutional weight of the baseline network, which involves both a learned binary mask and a few additional parameters. The binary mask 
is used as a scaled and shifted additive component and as a multiplicative filter to the original weights.  {Figure \ref{fig:teaser-masks} shows an example application of our algorithm. Given a network pretrained on a particular domain (\ie ImageNet \cite{russakovsky2015imagenet}, orange blocks) we can transform its original weights through binary masks (colored grids) and obtain a network which effectively addresses a novel domain (\eg digit \cite{netzer2011reading} or traffic sign \cite{stallkamp2012man} recognition)}.  
Our solution allows to achieve two main goals: 1) boosting the performance of each domain-specific network that we train, by leveraging the higher degree of freedom in {transforming} the baseline network, while 2) keeping a low per-domain overhead in terms of additional parameters (slightly more than 1 bit per parameter per domain).

We assess the validity of our method, and some variants thereof, on standard benchmarks including the Visual Decathlon Challenge~\cite{rebuffi2017learning}. The experimental results show that our model achieves performances comparable with fine-tuning separate networks for each recognition domain on all benchmarks while retaining a very small overhead in terms of additional parameters per domain. {Notably, we achieve results comparable to state-of-the-art models on the Visual Decathlon Challenge \cite{rebuffi2017learning} but without requiring multiple training stages \cite{Li_2019_CVPR} or a large number of domain-specific parameters \cite{Guo_2019_CVPR,rebuffi2018efficient}.}

{This paper extends our earlier work \cite{mancini2018adding} in many aspects. In particular, we provide a general formulation of binary mask based methods for multi-domain learning, with \cite{mancini2018adding} and \cite{mallya2018piggyback} obtained as special cases. We show how this general formulation allows boosting the performances of binary mask based methods in multiple scenarios, achieving close to state-of-the-art results in the Visual Domain Decathlon Challenge. Finally, we significantly expand our experimental evaluation by 1) considering more recent multi-domain learning methods, 2) ablating the various components of our model as well as various design choices and 3) showing additional quantitative and qualitative results, employing multiple backbone architectures.}


\section{Related works}\label{sec:related}
\textbf{Multi-domain Learning.}
{The need for visual models capable of addressing multiple domains received a lot of attention in recent years for what concerns both multi-task learning \cite{zamir2018taskonomy,liu2019end,cermelli2019rgb} and multi-domain learning \cite{rebuffi2017learning,rosenfeld2017incremental}. Multi-task learning focuses on learning multiple visual tasks (\eg semantic segmentation, depth estimation \cite{liu2019end}) with a single architecture. On the other hand, the goal of multi-domain learning is building a model able to address a task (e.g. classification) in multiple visual domains (e.g. real photos, digits) without forgetting previous domains and by using fewer parameters possible. An important work in this context is \cite{bilen2017universal}, where the authors showed how multi-domain learning can be addressed by using a network sharing all parameters except for batch-normalization (BN) layers \cite{ioffe2015batch}. In \cite{rebuffi2017learning}, the authors introduced the Visual Domain Decathlon Challenge, a first multi-domain learning benchmark. The first attempts in addressing this challenge involved domain-specific residual components added in standard residual blocks, either in series \cite{rebuffi2017learning} or in parallel \cite{rebuffi2018efficient}, 
In \cite{rosenfeld2017incremental} the authors propose to use controller modules where the parameters of the base architecture are recombined channel-wise, while in \cite{liu2019end} exploits domain-specific attention modules.  
Other effective approaches include devising instance-specific fine-tuning strategies \cite{Guo_2019_CVPR}, target-specific architectures \cite{Morgado_2019_CVPR} and learning covariance normalization layers \cite{Li_2019_CVPR}.

In \cite{mallya2017packnet} only a reserved subset of network parameters is considered for each domain.} The intersection of the parameters used by different domains is empty, thus the network can be trained end-to-end for each domain. Obviously, as the number of domain increases, fewer parameters are available for each domain, with a consequent limitation on the performances of the network. {To overcome this issue, in \cite{mallya2018piggyback} the authors proposed a more compact and effective solution based on directly learning domain-specific binary masks. 
The binary masks determine which of the network parameters are useful for the new domain and which are not, changing the actual composition of the features extracted by the network. This approach inspired subsequent works, improving both either the power of the binary masks \cite{mancini2018adding} or their amount of bits required, masking directly an entire channel \cite{berriel2019budget}. In this work, we take inspiration from these last research trends. In particular, we generalize the design of the binary masks employed in \cite{mallya2018piggyback} and \cite{mancini2018adding}. In particular, we consider neither simple multiplicative binary masks nor simple affine transformations of the original weights \cite{mancini2018adding} but a general and flexible formulation capturing both cases. 
Experiments show how our approach leads to a boost in the performances while using a comparable number of parameters per domain. Moreover, our approach achieves performances comparable to more complex models \cite{rebuffi2018efficient,Morgado_2019_CVPR,Li_2019_CVPR,Guo_2019_CVPR} in the challenging Visual Domain Decathlon challenge, largely reducing the gap of binary-mask based methods with the current state of the art. 
}

\vspace{8pt}\noindent\textbf{Incremental Learning.}
\tochange{The keen interest in incremental and life-long learning methods dates back to the pre-convnet era, with shallow learning approaches ranging from large margin classifiers \cite{KuzborskijOC13,KuzborskijOC17} to non-parametric methods \cite{MensinkVPC13,RistinGGG16}. 
In recent years, various works have addressed the problem of incremental and life-long learning  within the framework of deep architectures \cite{RebuffiKSL17,GuerrieroCM18,BendaleB16,cermelli2020modeling}. A major risk when training a neural network on a novel task/domain is to deteriorate the performances of the network on old domains, discarding previous knowledge. This phenomenon is called \textit{catastrophic forgetting} \cite{mccloskey1989catastrophic,french1999catastrophic,goodfellow2013empirical}. {To address this issue}, various works designed constrained optimization procedures taking into account the initial network weights, trained on previous domains. In \cite{li2017learning}, the authors exploit knowledge distillation \cite{hinton2015distilling} to obtain target objectives for previous domains/tasks, while training for novel ones. The additional objective ensures the preservation of the activation for previous domains, making the model less prone to experience the catastrophic forgetting problem. In \cite{kirkpatrick2017overcoming} the authors consider computing the update of the network parameters, based on their importance for previously seen domains. }
While these approaches are optimal in terms of the required parameters, i.e. they maintain the same number of parameters of the original network, they limit the catastrophic forgetting problem to the expenses of lower performance on both old and new domains. Recent methods overcome this issue by devising domain-specific parameters which are added as new domains are learned. If the initial network parameters remain untouched, the catastrophic forgetting problem is avoided but at the cost of the additional parameters required. The extreme case is the work of \cite{rusu2016progressive} in the context of reinforcement learning, where a parallel network is added each time a new domain is presented with side domain connections, exploited to improve the performances on novel domains.  

{Our work addresses catastrophic forgetting by adding domain-specific parameters, as in \cite{rusu2016progressive} and the mask-based approaches \cite{mallya2017packnet,mallya2018piggyback}. However, our domain-specific parameters require lower overhead with respect to \cite{rusu2016progressive} while at the same time being more effective than the ones in \cite{mallya2017packnet,mallya2018piggyback}.}

\vspace{8pt}\noindent\textbf{Network Binarization.}
Due to the low overhead required by the binary domain-specific parameters, our method is linked to recent works on binarization \cite{courbariaux2016binarized,hubara2016binarized,rastegari2016xnor} and quantization \cite{hubara2016quantized,lin2016fixed,zhou2016dorefa} of network parameters. {Binarization methods \cite{courbariaux2016binarized,hubara2016binarized} binarize network parameters and activations in order to obtain a lower computational cost, making the architecture usable in devices with constrained computational capabilities. Binarization can be performed at multiple levels. In \cite{courbariaux2016binarized,hubara2016binarized}, network weights and activations are binarized between -1 and 1 at run time and used to compute the parameters gradients. In \cite{rastegari2016xnor}, standard dot products are replaced by XNOR operations among binarized parameters and inputs. 
A closely related research thread is network quantization \cite{hubara2016quantized}, where instead of binary, low bitwidth networks and activations are considered. In \cite{zhou2016dorefa,lin2015neural}, also the network gradients are quantized, reducing the memory and computation footprint in the backward pass. } 

{Similarly to these works, here we are interested in obtaining a network representation with a low memory footprint. However, opposite to these works, we compress neither the whole architecture nor its activations but a subset of its parameters (i.e. the domain-specific ones) in order to make the sequential extension to multiple domains scalable. Despite these differences, the optimization techniques used in these works (\eg \cite{hubara2016binarized}) are fundamental building blocks of our algorithm.}

\section{Method}\label{sec:method}
{We address the problem of multi-domain 
learning, where we want to extend an architecture pretrained on a given task (\eg ResNet-50 pretrained on ImageNet) to perform the same task (\eg classification) in different domains (\eg traffic signal classification, digits recognition) with different output spaces.} {As in \cite{mallya2017packnet,mallya2018piggyback}, we consider the case where we receive the different domains one at the time, in a sequential fashion. We highlight that the term \textit{sequential} refers to the nature of the problem but our current formulation of the model extends the pretrained model one domain at the time, without considering the order in which the domains are received.}


In this context, our goal is to maximize the performance of the base model on the new set of domains, while limiting the memory occupied by the additional parameters needed. The solution we propose exploits the key idea from Piggyback~\cite{mallya2018piggyback} of learning domain-specific masks, but instead of pursuing the simple multiplicative transformation of the parameters of the baseline network, we define a parametrized, affine transformation mixing a binary mask and real parameters
that significantly increases the expressiveness of the approach, leading to a rich and nuanced ability to adapt the old parameters to the needs of the new domains. {This brings considerable improvements on the conducted experiments}, as we will show in the experimental section, while retaining a reduced, per-domain overhead.

\subsection{Overview}
{Let us assume to be given a pretrained, \emph{baseline} network $f_0(\cdot; \Theta, \Omega_0):\set X\to\set Y_0$ assigning a class label in $\set Y_0$ to elements of an input space $\set X$ (\eg images)}.\footnote{We focus on classification tasks, but the proposed method applies also to other tasks.}
{The parameters of the baseline network are partitioned into two sets: $\Theta$ comprises parameters that will be shared for other domains, whereas $\Omega_0$ entails the rest of the parameters (\eg the classifier).
Our goal is to learn for each domain $i\in\{1,\ldots,\con m\}$, with a possibly different output space $\set Y_i$,
a classifier $f_i(\cdot;\Theta,\Omega_i):\set X\to\set Y_i$. Here, $\Omega_i$ entails the parameters specific for the $i$th domain, while $\Theta$ holds the shareable parameters of the baseline network mentioned above.}
\begin{figure*}[t]
 \centering
 \includegraphics[width=0.8\textwidth]{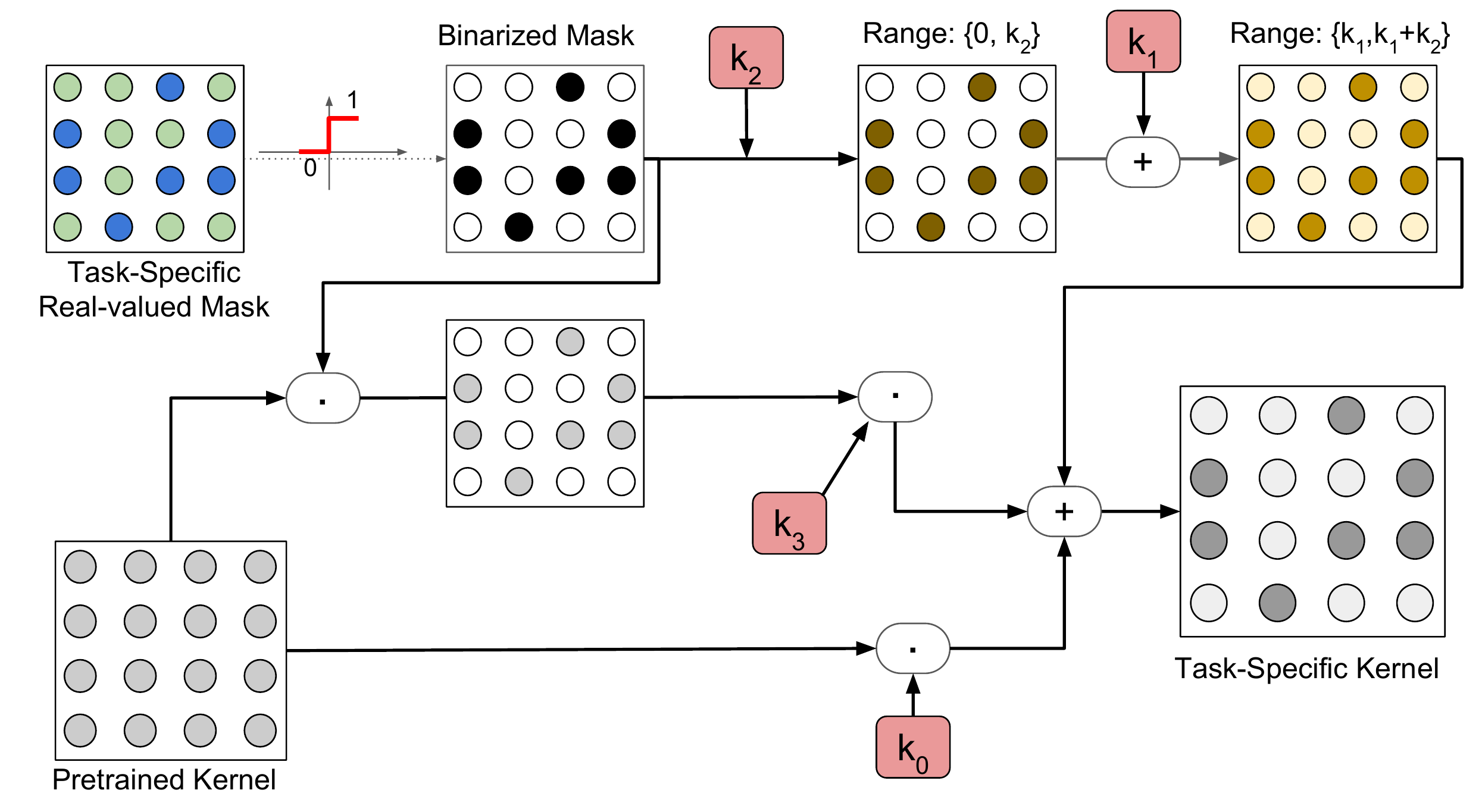}
     \caption{Overview of the proposed model (best viewed in color). Given a convolutional kernel, for each domain, we exploit a real-valued mask to generate a domain-specific binary mask. 
An affine transformation directly applied to the binary masks, which changes their range (through a scale parameter $k_2$) and their minimum value (through $k_1$). A multiplicative mask applied to the original kernels and the pretrained kernel themselves are scaled by the factors $k_3$ and $k_0$ respectively. All the different masks are summed to produce the final domain-specific kernel. }
    \label{fig:method-full}
 \end{figure*}

 Each domain-specific network $f_i$ shares the same structure of the baseline network $f_0$, except for having a possibly differently sized classification layer.
{ For each convolutional layer\footnote{Fully-connected layers are a special case.} of $f_0$ with parameters $\mat W$, the domain-specific network $f_i$ holds a binary mask $\mat M$, with the same shape of $\mat W$, that is used to mask original filters. The way the mask is exploited to specialize the network filters produces different variants of our model, which we describe in the following.} 

\subsection{Affine Weight Transformation through Binary Masks}
{Following previous works \cite{mallya2018piggyback,mancini2018adding}, we consider domain-specific networks $f_i$ that are shaped as the baseline network $f_0$ and we store in $\Omega_i$ a binary mask $\mat M$ for each convolutional kernel $\mat W$ in the shared set $\Theta$. 
However, differently from \cite{mallya2018piggyback,mancini2018adding}, we consider a more general affine transformation of the base convolutional kernel $\mat W$ that depends on a binary mask $\mat M$ as well as additional parameters. Specifically, we transform $\mat W$ into}
\begin{equation}\label{eq:ours}
\tilde {\mat W}=k_0\mat W+k_1 \mathtt 1+k_2\mat M +k_3\mat W\circ\mat M\,,
\end{equation}
where $k_j\in\mathbb R$ are additional domain-specific parameters in $\Omega_i$ that we learn along with the binary mask $\mat M$, $\mat 1$ is an opportunely sized tensor of $1$s, and $\circ$ is the Hadamard (or element-wise) product.  The transformed parameters $\hat{\mat W}$ are then used in the convolutional layer of $f_i$. 
 {We highlight that the domain-specific parameters that are stored in $\Omega_i$ amount to just a single bit per parameter in each convolutional layer plus a few scalars per layer, yielding a low overhead per additional domain while retaining a sufficient degree of freedom to build new convolutional weights. Figure~\ref{fig:method-full} provides an overview of the transformation in \eqref{eq:ours}.
 
Our model, can be regarded as a parametrized generalization of~\cite{mallya2018piggyback}, since we can recover the formulation of~\cite{mallya2018piggyback} by setting $k_{0,1,2}=0$ and $k_3=1$. }
Similarly, if we get rid of the multiplicative component, \ie we set $k_3=0$, we obtained the following simplified transformation
\begin{equation}\label{eq:simple}
\check {\mat W}=k_0\mat W+k_1 \mathtt 1+k_2\mat M\,,
\end{equation}
which corresponds to the method presented in our previous work \cite{mancini2018adding} and will be taken into account in our analysis.

{We want to highlight that each model (i.e. ours,\cite{mallya2018piggyback} and \cite{mancini2018adding}) has different representation capabilities. In fact, in \cite{mallya2018piggyback}, the domain-specific parameters can take only two possible values: either $0$ (i.e. if $m=0$) or the original pretrained weights (i.e. if $m=1$). On the other hand, the scalar components of our previous work \cite{mancini2018adding} allow both scaling (i.e. with $k_0$) and shifting (i.e. with $k_1$) the original network weights, with the additive binary mask adding a bias term (i.e. $k_2$) selectively to a group of parameters (\ie the one with $m=1$). Our work generalizes \cite{mallya2018piggyback} and \cite{mancini2018adding} by considering the multiplicative binary-mask term $\mat W \circ \mat M$ as an additional bias component scaled by the scalar $k_3$. In this way, our model has the possibility to obtain \textit{parameter-specific} bias components, something that was not possible neither in \cite{mallya2018piggyback} nor in \cite{mancini2018adding}. The additional degrees of freedom makes the search space of our method larger with respect to \cite{mallya2018piggyback,mancini2018adding}, with the possibility to express more complex (and tailored) domain-specific transformations. Thus, as we show in the experimental section, the additional parameters that we introduce with our method bring a negligible per-domain overhead compared to \cite{mallya2018piggyback} and \cite{mancini2018adding}, which is nevertheless generously balanced out by a significant boost of the performance of the domain-specific classifiers.}

{Finally, following \cite{bilen2017universal,mancini2018adding}, we opt also for domain-specific batch-normalization parameters (\ie mean, variance, scale and bias), unless otherwise stated. 
Those parameters will not be fixed (i.e. they do not belong to $\Theta$) but are part of $\Omega_i$, and thus optimized for each domain. 
As in \cite{mancini2018adding}, in the cases where we have a convolutional layer followed by batch normalization, we keep the corresponding parameter $k_0$ fixed to $1$, because the output of batch normalization is invariant to the scale of the convolutional weights.
}

\subsection{Learning Binary Masks}
{Given the training set of the $i_{th}$ domain, we learn the domain-specific parameters $\Omega_i$ 
by 
minimizing a standard supervised loss, \ie the classification log-loss. However, while the domain-specific batch-normalization parameters can be learned by employing standard stochastic optimization methods, the same is not feasible for the binary masks. Indeed, optimizing the binary masks directly would turn the learning into a combinatorial problem. {To address} this issue, we follow the solution adopted in~\cite{mallya2018piggyback,mancini2018adding}, \ie we
replace each binary mask $\mat M$ with a thresholded real matrix $\mat R$. By doing so, we shift from optimizing discrete variables in $\mat M$ to continuous ones in $\mat R$. However, the gradient of the hard threshold function $h(r)=1_{r\geq 0}$ is zero almost everywhere, making this solution apparently incompatible with gradient-based optimization approaches. To address this issue we consider a strictly increasing, surrogate function $\tilde h$ that will be used in place of $h$ \emph{only} for the gradient computation, \ie \[
h'(r)\approx \tilde h'(r)\,,
\]
where $h'$ denotes the derivative of $h$ with respect to its argument.
The gradient that we obtain via the surrogate function has the property that it always points in the right down hill direction in the error surface.} {Let $r$ be a single entry of $\mat R$, with $m=h(r)$} and let $E(m)$ be the error function. Then 
\[
\sign((E\circ h)'(r))=\sign(E'(m)h'(r))=\sign\left(E'(m)\tilde h'(r)\right)
\]
and, since $\tilde h'(r)>0$ by construction of $\tilde h$, we obtain the sign agreement
\[
\sign\left((E\circ h)'(r)\right)=\sign\left(E'(m)\right)\,.
\]
Accordingly, when the gradient of $E(h(r))$ with respect to $r$ is positive (negative), this induces a decrease (increase) of $r$. By the monotonicity of $h$ this eventually induces a decrease (increase) of $m$, which is compatible with the direction pointed by the gradient of $E$ with respect to $m$.

{In the experiments, we set $\tilde h(x)=x$, \ie the identity function, recovering the workaround suggested in~\cite{hin12} and employed also in~\cite{mallya2018piggyback,mancini2018adding}. 
However, other choices are possible. For instance, by taking $\tilde h(x)=(1+e^{-x})^{-1}$, \ie the sigmoid function, we obtain a better approximation that has been suggested in~\cite{goodman1994learning,bengio2013estimating}}. We test different choices for $\tilde h(x)$ in the experimental section.

\section{Experiments}\label{sec:experiments}
\vspace{-10pt}
\myparagraph{Datasets.}
{In the following, we test our method on two different multi-domain benchmarks, where the multiple domains regard different classification tasks}. For the first benchmark we follow \cite{mallya2018piggyback}, and we use 6 datasets: ImageNet \cite{russakovsky2015imagenet}, VGG-Flowers \cite{nilsback2008automated}, Stanford Cars \cite{krause20133d}, Caltech-UCSD Birds (CUBS) \cite{wah2011caltech}, Sketches \cite{eitz2012humans} and WikiArt \cite{saleh2015large}. VGG-Flowers \cite{nilsback2008automated} is a dataset of fine-grained recognition containing images of 102  categories, corresponding to different kind of flowers. There are 2'040 images for training and 6'149 for testing. Stanford Cars \cite{krause20133d} contains images of 196 different types of cars with approximately 8 thousand images for training and 8 thousand for testing. Caltech-UCSD Birds \cite{wah2011caltech} is another dataset of fine-grained recognition containing images of 200 different species of birds, with approximately 6 thousand images for training and 6 thousand for testing. Sketches \cite{eitz2012humans} is a dataset composed of 20 thousand sketch drawings, 16 thousand for training and 4 thousand for testing. It contains images of 250 different objects in their sketched representations. WikiArt \cite{saleh2015large} contains painting from 195 different artists. The dataset has 42'129 images for training and 10628 images for testing. These datasets contain a lot of variations both from the category addressed (\ie cars \cite{krause20133d} vs birds \cite{wah2011caltech}) and the appearance of their instances (from natural images \cite{russakovsky2015imagenet} to paintings \cite{saleh2015large} and sketches \cite{eitz2012humans}), thus representing a challenging benchmark for multi-domain learning techniques. 

The second benchmark 
is the Visual Domain Decathlon Challenge \cite{rebuffi2017learning}. This challenge has been introduced to check the capability of a single algorithm to tackle 10 different classification tasks. The tasks are taken from the following datasets: ImageNet \cite{russakovsky2015imagenet}, CIFAR-100 \cite{krizhevsky2009learning}, Aircraft \cite{maji2013fine}, Daimler pedestrian classification (DPed) \cite{munder2006experimental}, Describable textures (DTD) \cite{cimpoi2014describing}, German traffic signs (GTSR) \cite{stallkamp2012man} , Omniglot \cite{lake2015human}, SVHN \cite{netzer2011reading}, UCF101 Dynamic Images \cite{bilen2016dynamic,soomro2012ucf101} and VGG-Flowers \cite{nilsback2008automated}. A more detailed description of the challenge and the datasets can be found in \cite{rebuffi2017learning}. For this challenge, an independent scoring function is defined \cite{rebuffi2017learning}. This function $S$ is expressed as:
\begin{equation}\label{eq:S}
S=\sum^{10}_{d=1}\alpha_d\text{max}\{0,E_d^{\text{max}}-E_d\}^{2}
\end{equation}
where $E_d^{\text{max}}$ is the test error of the baseline in the domain $d$, $E_d$ is the test error of the submitted model and $\alpha$ is a scaling parameter ensuring that the perfect score for each domain is 1000, thus with a maximum score of 10000 for the whole challenge. The baseline error is computed doubling the error of 10 independent models fine-tuned on the single domains.
This score function takes into account the performances of a model on all 10 classes, preferring models with good performances on all of them compared to models outperforming by a large margin the baseline in just a few. {Following \cite{berriel2019budget}, we use this metric also for the first benchmark, keeping the same upper-bound of 1000 points for each domain. Moreover, as in \cite{berriel2019budget}, we report the ratio among the score obtained and the parameters used, denoting it as ${S}_p$. This metric allows capturing the trade-off among the performances and model size.}

\myparagraph{Networks and training protocols.} For the first benchmark, we use 3 networks: ResNet-50 \cite{he2016deep},  DenseNet-121 \cite{huang2017densely} and VGG-16 \cite{simonyan2014very}, reporting the results of Piggyback \cite{mallya2018piggyback}, PackNet \cite{mallya2017packnet} and both the simple \cite{mancini2018adding} and full version of our model described in Section~\ref{sec:method}. 

Following the protocol of \cite{mallya2018piggyback}, for all the models we start from the networks pretrained on ImageNet and train the domain-specific networks using Adam \cite{kingma2014adam} as optimizer except for the classifiers where SGD \cite{bottou2010large} with momentum is used. The networks are trained with a batch-size of 32 and an initial learning rate of 0.0001 for Adam and 0.001 for SGD with momentum 0.9. Both the learning rates are decayed by a factor of 10 after 15 epochs. In this scenario, we use input images of size $224\times224$ pixels, with the same data augmentation (\ie mirroring and random rescaling) of \cite{mallya2017packnet,mallya2018piggyback}. The real-valued masks are initialized with random values drawn from a uniform distribution with values between $0.0001$ and $0.0002$. Since our model is independent of the order of the domains, we do not take into account different possible orders, reporting the results as accuracy averaged across multiple runs. For simplicity, in the following, we will denote this scenario as \textit{ImageNet-to-Sketch}.

For the Visual Domain Decathlon, we employ the Wide ResNet-28 \cite{zagoruyko2016wide} adopted by previous methods \cite{rebuffi2017learning,rosenfeld2017incremental,mallya2018piggyback}, with a widening factor of 4 (\ie 64, 128 and 256 channels in each residual block). Following \cite{rebuffi2017learning} we rescale the input images to $72\times72$ pixels giving as input to the network images cropped to $64\times64$. We follow the protocol in \cite{mallya2018piggyback}, by training the simple and full versions of our model for 60 epochs for each domain, with a batch-size of 32, and using again Adam for the entire architecture but the classifier, where SGD with momentum is used. The same learning rates of the first benchmark are adopted and are decayed by a factor of 10 after 45 epochs. Similarly, the same initialization scheme is used for the real-valued masks. \textit{No hyperparameter tuning} has been performed as we used a single training schedule for all the 10 domains, except for the ImageNet pretrained model, which was trained following the schedule of \cite{rebuffi2017learning}. As for data augmentation, mirroring has been performed, except for the datasets with digits (\ie SVHN), signs (Omniglot, GTSR) and textures (\ie DTD) as it may be rather harmful (as in the first 2 cases) or unnecessary.

{In both benchmarks, we train our network on one domain at the time, sequentially for all domains. For each domain, we introduce the domain-specific binary masks and additional scalar parameters, as described in section \ref{sec:method}. Moreover, following previous approaches \cite{rebuffi2017learning,rebuffi2018efficient,mallya2018piggyback,rosenfeld2017incremental}, we consider a separate classification layer for each domain. This is reflected also in the computation of the parameters overhead required by our model, we do not consider the separate classification layers, following comparison systems \cite{rebuffi2017learning,rebuffi2018efficient,mallya2018piggyback,rosenfeld2017incremental}.}

\subsection{Results}
\vspace{-10pt}
\myparagraph{ImageNet-to-Sketch.} In the following, we discuss the results obtained by our model on the ImageNet-to-Sketch scenario. We compare our method with Piggyback \cite{mallya2018piggyback}, PackNet \cite{mallya2017packnet} and two baselines considering (i) the network only as feature extractor, training only the domain-specific classifier, and (ii) individual networks separately fine-tuned on each domain. PackNet \cite{mallya2017packnet} adds a new domain to a pretrained architecture by identifying which weights are important for the domain, optimizing the architecture through alternated pruning and re-training steps. Since this algorithm is dependent on the order of the domains, we report the performances for two different orderings \cite{mallya2018piggyback}: starting from the model pretrained on ImageNet, in the first setting ($\downarrow$) the order is CUBS-Cars-Flowers-WikiArt-Sketch while for the second ($\uparrow$) the order is reversed. For our model, we evaluate both the full and the simple version, including domain-specific batch-normalization layers. Since including batch-normalization layers affects the performances, for the sake of presenting a fair comparison, we report also the results of Piggyback \cite{mallya2018piggyback} obtained as a special case of our model with separate BN parameters per domain for ResNet-50 and DenseNet-121. {Moreover, we report the results of the Budget-Aware adapters ($\text{BA}^2$) method in \cite{berriel2019budget}. This method relies on binary masks applied not per-parameter but per-channel, with a budget constraint allowing to further squeeze the network complexity. As in our method, also in \cite{berriel2019budget} domain-specific BN layers are used.}

Results are reported in Tables \ref{tab:resnet-ImageNet}, \ref{tab:densenet-ImageNet} and \ref{tab:vgg-ImageNet}. We see that both versions of our model are able to fill the gap between the classifier only baseline and the individual fine-tuned architectures almost entirely and in all settings. For larger and more diverse datasets such as Sketch and WikiArt, the gap is not completely covered, but the distance between our models and the individual architectures is always less than 1\%. These results are remarkable given the simplicity of our method, not involving any assumption of the optimal weights per domain \cite{mallya2017packnet}, and the small overhead in terms of parameters that we report in the row "\#~Params" (\ie $1.17$ for ResNet-50, $1.21$ for DenseNet-121 and $1.16$ for VGG-16), which represents the total number of parameters (counting all domains and excluding the classifiers) relative to the ones in the baseline network\footnote{If the base architecture contains $N_p$ parameters and the additional bits introduced per domain are $A_p$ then ${\text{\#~Params}=1+\frac{A_p\cdot (T-1)}{32\cdot N_p}}$,
where $T$ denotes the number of domains (included the one used for pretraining the network) and the 32 factor comes from the bits required for each real number. The classifiers are not included in the computation.}.
For what concerns the comparison with the other algorithms, our model consistently outperforms both the basic version of Piggyback and PackNet in all the settings and architectures, except Sketch for the DenseNet and VGG-16 architectures and CUBS for VGG-16, in which the performances are comparable with those of Piggyback. When domain-specific BN parameters are introduced also for Piggyback (Tables \ref{tab:resnet-ImageNet} and \ref{tab:densenet-ImageNet}), the gap in performances is reduced, with performances comparable to those of our model in some settings (\ie CUBS) but with still large gaps in others (\ie Flowers, Stanford Cars and WikiArt). These results show that the advantages of our model are not only due to the additional BN parameters, but also to the more flexible and powerful affine transformation introduced. This statement is further confirmed with the VGG-16 experiments in Table \ref{tab:vgg-ImageNet}.  For this network, when the standard Piggyback model is already able to fill the gap between the feature extractor baseline and the individual architectures, our model achieves either comparable or slightly superior performances (\ie CUBS, WikiArt and Sketch). However, in the scenarios where Piggyback does not reach the performances of the independently fine-tuned models (\ie Stanford Cars and Flowers), our model consistently outperforms the baseline, either halving (Flowers) or removing (Stanford Cars) the remained gap. Since this network does not contain batch-normalization layers, it confirms the generality of our model, showing the advantages of both our simple and full versions, even without domain-specific BN layers. 

{For what concerns the comparison with $\text{BA}^2$, the performances of our model are either comparable or superior in most of the settings. Remarkable are the gaps in the WikiArt dataset, with our full model surpassing $\text{BA}^2$ by 3\% with ResNet-50 and 4\% for DenseNet-121. Despite both Piggyback and $\text{BA}^2$ use fewer parameters than our approach, our full model outperforms both of them in terms of the final score (Score row) and the ratio among the score and the parameters used (Score/Params row). This shows that our model is the most powerful in making use of the binary masks, achieving not only higher performances but also a more favorable trade-off between performances and model size.} 

Finally, both Piggyback, $\text{BA}^2$ and our model outperform PackNet and, as opposed to the latter method, do not suffer from the heavy dependence on the ordering of the domains. This advantage stems from having a multi-domain learning strategy that is domain-independent, with the base network not affected by the new domains that are learned.

\begin{table*}[t]
			\caption{Accuracy of ResNet-50 architectures in the ImageNet-to-Sketch scenario.} 
                   
		\centering
        \scalebox{1}{
		\begin{tabular}{ l | c || ec | ec | ec | ec | ec | ec | ec || c  } 
			\multirow{2}{*}{Dataset} & Classifier & \multicolumn{2}{ c | }{PackNet\cite{mallya2018piggyback}} &\multicolumn{2}{c |}{Piggyback}& $\text{BA}^2$ & \multicolumn{2}{ c ||}{Ours} & Individual\\
          & Only \cite{mallya2018piggyback} &$\downarrow$&$\uparrow$ & \cite{mallya2018piggyback}&BN& \cite{berriel2019budget}&Simple& Full  & \cite{mallya2018piggyback}\\ \hline
             \# Params&1&\multicolumn{2}{c |}{1.10}&1.16&1.17& 1.03&1.17&1.17&6\\  \hline
       ImageNet &76.2&75.7&75.7&\textbf{76.2} &\textbf{76.2}&\textbf{76.2}&\textbf{76.2}&\textbf{76.2} &76.2\\
       CUBS&70.7 &80.4&71.4&80.4 &82.1&81.2&\textbf{82.6} &82.4&82.8 \\
       Stanford Cars&52.8&86.1&80.0 &88.1 &90.6&\textbf{92.1}&{91.5}&91.4&91.8 \\
       Flowers&86.0&93.0&90.6&93.5&95.2&95.7&96.5&\textbf{96.7} &96.6\\
       WikiArt&55.6&69.4&70.3&73.4 &74.1&72.3&74.8&\textbf{75.3}&75.6\\
       Sketch&50.9&76.2&78.7&79.4 &79.4&79.3&\textbf{80.2}&\textbf{80.2} &80.8 \\
            \hline
        Score&533&732&620&934&1184&1265&1430&\textbf{1458}&1500\\
        Score/Params&533&665&534&805&1012&1228&1222&\textbf{1246}&250\\
		\end{tabular}
        }
		\label{tab:resnet-ImageNet}
\end{table*}
\begin{table*}[t]
			\caption{Accuracy of DenseNet-121 architectures in the ImageNet-to-Sketch scenario.} 
\centering
		\begin{tabular}{ l | c || ec | ec| ec | ec | ec | ec | ec || c  } 
			\multirow{2}{*}{Dataset} & Classifier &  \multicolumn{2}{c |}{PackNet\cite{mallya2018piggyback}}&\multicolumn{2}{c |}{Piggyback} & $\text{BA}^2$ &\multicolumn{2}{c ||}{Ours} & Individual\\
          & Only \cite{mallya2018piggyback}&$\downarrow$ & $\uparrow$&\cite{mallya2018piggyback} &BN&\cite{berriel2019budget}&Simple&Full & \cite{mallya2018piggyback}\\ 
            \hline
           \# Params&1&\multicolumn{2}{c |}{1.11}&1.15& 1.21 &1.17&1.21&1.21&6 \\  \hline
       ImageNet &74.4 &\textbf{74.4}&\textbf{74.4}&\textbf{74.4} &\textbf{74.4}&\textbf{74.4} & \textbf{74.4}& \textbf{74.4}&74.4\\
       CUBS &73.5 &80.7&69.6&79.7 &81.4&\textbf{82.4}&81.5&{81.7}&81.9\\
       Stanford Cars&56.8 &84.7&77.9&87.2 &90.1&\textbf{92.9}&{91.7} &91.6&91.4\\
       Flowers&83.4 &91.1&91.5&94.3 &95.5&96.0&96.7&\textbf{96.9} &96.5\\
       WikiArt&54.9 &66.3&69.2 &72.0&73.9&71.5&75.5&\textbf{75.7} &76.4\\
       Sketch&53.1 &74.7&78.9&\textbf{80.0} &79.1&79.9&79.9&79.8 &80.5 \\
       \hline
        Score&324&685&607&946&1209&1434&1506&\textbf{1534}&1500\\
        Score/Params&324&617&547&822&999&1226&1245&\textbf{1268}&250\\
		\end{tabular}
		\label{tab:densenet-ImageNet}
        \vspace{-5pt}
\end{table*}

\begin{table*}[t]
			\caption{Accuracy of VGG-16 architectures in the ImageNet-to-Sketch scenario.} 
            
\vspace{5pt}            
		\centering
        \scalebox{1}{
		\begin{tabular}{ l | c || ec | ec | c | ec | ec || c  } 
			\multirow{2}{*}{Dataset} & Classifier & \multicolumn{2}{ c | }{PackNet\cite{mallya2018piggyback}} &Piggyback & \multicolumn{2}{ c ||}{Ours} & Individual\\
          & Only \cite{mallya2018piggyback} &$\downarrow$&$\uparrow$ & \cite{mallya2018piggyback}& Simple & Full  & \cite{mallya2018piggyback}\\ \hline
             \# Params&1&\multicolumn{2}{c |}{1.09}&1.16&1.16&1.16&6\\\hline
       ImageNet &71.6&70.7&70.7&\textbf{71.6} &\textbf{71.6}&\textbf{71.6} &71.6\\
       CUBS&63.5 &77.7&70.3&\textbf{77.8} &77.4 &77.4&77.4 \\
       Stanford Cars&45.3&84.2&78.3 &86.1&87.2&\textbf{87.3}&87.0 \\
       Flowers&80.6&89.7&89.8&90.7&\textbf{91.6}&91.5&92.3\\
       WikiArt&50.5&67.2&68.5&71.2 &71.6&\textbf{71.9}&67.7\\
       Sketch&41.5&71.4&75.1&76.5&76.5&\textbf{76.7 }&76.4 \\
            \hline
        Score&342&1152&979&1441&1530&\textbf{1538}&1500\\
        Score/Params&342&1057&898&1243&1319&\textbf{1326}&250\\
		\end{tabular}
        }
		\label{tab:vgg-ImageNet}
\end{table*}

\myparagraph{Visual Decathlon Challenge.}
In this section, we report the results obtained on the Visual Decathlon Challenge. {We compare our model with the baseline method Piggyback \cite{mallya2018piggyback} (PB), the budget-aware adapters of \cite{berriel2019budget} ($\text{BA}^2$), the improved version of the winning entry of the 2017 edition of the challenge \cite{rosenfeld2017incremental} (DAN), the network with domain-specific parallel adapters \cite{rebuffi2018efficient} (PA), the domain-specific attention modules of \cite{liu2019end} (MTAN), the covariance normalization approach \cite{Li_2019_CVPR} (CovNorm) and SpotTune \cite{Guo_2019_CVPR}}. We additionally report the baselines proposed by the authors of the challenge \cite{rebuffi2017learning}. For the latter, we report the results of 5 models: the network used as feature extractor (Feature), 10 different models fine-tuned on the single domains (Finetune), the network with domain-specific residual adapter modules \cite{rebuffi2017learning} (RA), the same model with increased weight decay (RA-decay) and the same architecture jointly trained on all 10 domains, in a round-robin fashion (RA-joint). The first two models are considered as references. For the parallel adapters approach \cite{rebuffi2018efficient} we report also the version with a post-training low-rank decomposition of the adapters (PA-SVD). This approach extracts a domain-specific and a domain agnostic component from the learned adapters with the domain-specific components which are further fine-tuned on each domain. Additionally, we report the novel results of the residual adapters \cite{rebuffi2017learning} as reported in \cite{rebuffi2018efficient} (RA-N).

Similarly to \cite{rosenfeld2017incremental} we tune the training schedule, jointly for the 10 domains, using the validation set, and evaluate the results obtained on the test set (via the challenge evaluation server) by a model trained on the union of the training and validation sets, using the validated schedule. As opposed to methods like~\cite{rebuffi2017learning} we use the same schedule for the 9 domains (except for the baseline pretrained on ImageNet), without adopting domain-specific strategies for setting the hyper-parameters. Moreover, we do not employ our algorithm while pretraining the ImageNet architecture as in~\cite{rebuffi2017learning}. For fairness, we additionally report the results obtained by our implementation of \cite{mallya2018piggyback} using the same pretrained model, training schedule and data augmentation adopted for our algorithm (PB ours).

The results are reported in Table 
\ref{tab:vdc-accuracy} 
{in terms of the $S$-score (see, Eq. \eqref{eq:S}) and $S_p$}. In the first part of the table are shown the baselines (\ie finetuned architectures and using the base network as feature extractor) while in the middle the models considering a sequential formulation of the problem, against which we compare. In the last part of the table we report, for fairness, the methods that do not consider a sequential multi-domain learning scenario since they either train on all the datasets jointly (RA-joint) or have a multi-process step considering all domains (PA-SVD).

From the table we can see that the full form of our model (F) {achieves very high results, being the third best performing method in terms of $S$-score, behind only CovNorm  and SpotTune and being comparable to PA. However, SpotTune uses a large amount of parameters (11x) and PA doubles the parameters of the original model. CovNorm uses a very low number of parameters but requires a two-stage pipeline. On the other hand, our model requires neither a large number of parameters (such as SpotTune and PA) nor a two-stage pipeline (as CovNorm) while achieving results close to the state of the art (215 points below CovNorm in terms of $S$-score). Compared to binary mask based approaches, our model surpasses PiggyBack of more than 600 points, $\text{BA}^2$ of 300 and the simple affine transformation presented in \cite{mancini2018adding} of more than 200. It is worth highlighting that these results have been achieved \textit{without domain-specific hyperparameter tuning}, differently from previous works e.g. \cite{rebuffi2017learning,rebuffi2018efficient,Li_2019_CVPR}.

For what concerns the $S_p$ score, our model is the third-best performing model, behind $\text{BA}^2$ and CovNorm. We highlight however that CovNorm requires a two-stage pipeline to reduce the amount of parameters needed, while $\text{BA}^2$ is explicitly designed with the purpose of limiting the budget (i.e. parameters, flops) required by the model.} 

\begin{table*}[t]
			\caption{Results in terms of $S$ and $S_p$ scores for the Visual Decathlon Challenge.} 
            
		\centering
        \scalebox{.95}{
		\begin{tabular}{ l | c || c  c  c  c  c  c  c  c  c  c | c c } 
			 Method&\#Params&ImNet&Airc. &C100&DPed&DTD&GTSR&Flwr.&Oglt.&SVHN&UCF& Score & ${S}_p$\\\hline
            Feature \cite{rebuffi2017learning}&1&59.7	&23.3	&63.1	&80.3	&45.4	&68.2	&73.7	&58.8	&43.5	&26.8	&544&544\\
            Finetune \cite{rebuffi2017learning}&10&59.9	&60.3	&82.1	&92.8	&55.5	&97.5	&81.4	&87.7	&96.6	&51.2	&2500&250\\
      \hline     RA\cite{rebuffi2017learning}&2&59.7	&56.7	&81.2	&93.9	&50.9	&97.1	&66.2	&{89.6}	&96.1	&47.5	&2118&1059\\
           RA-decay\cite{rebuffi2017learning}&2&59.7	&61.9	&81.2	&93.9	&57.1	&97.6	&81.7	&{89.6}	&96.1	&{50.1}	&2621&1311\\
           RA-N\cite{rebuffi2018efficient}&2&{60.3}&61.9&81.2&93.9&57.1&{99.3}&81.7&{89.6}&96.6&{50.1}&3159&1580\\
            DAN \cite{rosenfeld2017incremental}&2.17&57.7	&64.1	&80.1	&91.3	&56.5	&98.5	&86.1	&{89.7}	&{96.8}	&49.4	&2852&1314\\
             PA \cite{rebuffi2018efficient}&2&{60.3}&{64.2}&81.9&94.7&58.8&{99.4}&84.7&89.2&96.5&{50.9}&{3412}&1706\\
            MTAN \cite{liu2019end}&1.74&{63.9}&{61.8}&81.6&91.6&56.4&{98.8}&81.0&89.8&96.9&{50.6}&{2941}&1690\\
            SpotTune \cite{Guo_2019_CVPR} & 11 & 60.3 &63.9 &80.5&96.5&57.1&99.5&85.2&88.8&96.7&52.3&{3612}&328\\
            CovNorm \cite{Li_2019_CVPR} &1.25&60.4&69.4&81.3&98.8&60.0&99.1&83.4&87.7&96.6&48.9&\textbf{3713}&{2970}\\
            PB \cite{mallya2018piggyback}&1.28&57.7	&{65.3}	&79.9	&{97.0}	&57.5	&97.3	&79.1	&87.6	&{97.2}	&47.5	&2838&2217\\
            PB ours &1.28&{60.8}&52.3&80.0&95.1&{59.6}&98.7&82.9&85.1&96.7&46.9&2805&2191\\
            $\text{BA}^2$ \cite{berriel2019budget}&1.03&56.9	&49.4&78.1&95.5&55.1&99.4&86.1&88.7&96.9&50.2&3199&\textbf{3106}\\
          	Ours (S) \cite{mancini2018adding}&1.29&{60.8}	&51.3	&{81.9}	&94.7	&{59.0}	&99.1	&{88.0}	&89.3	&96.5	&48.7	&3263&2529\\
            Ours (F)&1.29&{60.8}	&52.8	&{82.0}	&{96.2}	&58.7	&{99.2}	&{88.2}	&89.2	&{96.8}	&48.6	&{3497}&{2711}\\
            \hline
            PA-SVD\cite{rebuffi2018efficient}&1.5&60.3&66.0&81.9&94.2&57.8&99.2&85.7&89.3&96.6&52.5&3398&2265\\
            RA-joint\cite{rebuffi2017learning}&2&59.2	&63.7	&81.3	&93.3	&57.0	&97.5	&83.4	&89.8	&96.2	&50.3	&2643&1322\\
            \hline
             
		\end{tabular}}
		\label{tab:vdc-accuracy}
        \vspace{-10pt}
\end{table*}

\subsection{Ablation Study}
\label{sec:ablation}

In the following, we analyze the impact of the various components of our model. In particular, we consider the impact of the parameters $k_0$, $k_1$, $k_2$, $k_3$ and the surrogate function $\tilde{h}$ on the final results of our model for the ResNet-50 and DenseNet-121 architectures in the ImageNet-to-Sketch scenario. Since the architectures contain batch-normalization layers, we set $k_0=1$ for our simple\cite{mancini2018adding} and full versions and $k_0=0$ when we analyze the special case \cite{mallya2018piggyback}. For the other parameters we adopt various choices: either we fix them to a constant to not take into account their impact, or we train them, to assess their particular contribution to the model. The surrogate function we use is the identity function $\tilde{h}(x)=x$ unless otherwise stated (\ie \textit{with Sigmoid}). The results of our analysis are shown in Tables \ref{tab:ablation-res} and \ref{tab:ablation-dense}. 

As the Tables show, while the BN parameters allow a boost in the performances of Piggyback, adding $k_1$ to the model does not provide a further gain in performances. This does not happen for the simple version of our model: without $k_1$ our model is not able to fully exploit the presence of the binary masks, achieving comparable or even lower performances with respect to the Piggyback model. We also note that a similar drop affecting our \emph{Simple} version \cite{mancini2018adding} when bias was omitted.  

Noticeable, the full versions with $k_2=0$ suffer a large decrease in performances in almost all settings (\eg ResNet-50 Flowers from 96.7\% to 91.0\%), showing that the component that brings the largest benefits to our algorithm is the addition of the binary mask itself scaled by $k_2$ (\ie $k_2 \cdot \mat M$). This explains also the reason why the simple version achieves performance similar to the full version of our model. We finally note that there is a limited contribution brought by the standard Piggyback component (\ie $k_1\cdot \mat W\circ \mat M$), compared to the new components that we have introduced in the transformation:
in fact, there is a clear drop in performance in various scenarios (\eg CUBS, Cars) when we set either $k_1=0$ or $k_2=0$. 
Consequently, as $k_1$ is introduced in our \emph{Simple} model, the boost of performances is significant such that neither the inclusion of $k_3$, nor considering channel-wise parameters $k_1$ provides further gains. Slightly better results are achieved in larger datasets, such as WikiArt, with the additional parameters giving more capacity to the model, thus better handling the larger amount of information available in the dataset. 

As to what concerns the choice of the surrogate $\tilde{h}$, no particular advantage has been noted when $\tilde{h}(x)=\sigma(x)$ with respect to the standard straight-through estimator ($\tilde{h}(x)=x$). This may be caused by the noisy nature of the straight-through estimator, which has the positive effect of regularizing the parameters, as shown in previous works \cite{bengio2013estimating,neelakantan2015adding}.

We also note that for DenseNet-121, as opposed to ResNet-50, setting $k_1$ to zero degrades the performance only in 1 out of 5 datasets (\ie CUBS) while the other 4 are not affected, showing that the effectiveness of different components of the model is also dependent on the architecture used.

\begin{table*}[t]
			\caption{Impact of the parameters $k_0$, $k_1$, $k_2$ and $k_3$ of our model using the ResNet-50 architectures in the ImageNet-to-Sketch scenario. \ding{51} denotes a learned parameter, while~$^*$ denotes \cite{mallya2018piggyback} obtained as a special case of our model.} 

		\vspace{5pt}
		\centering
		\begin{tabular}{ l | c | c | c | c | c | c | c | c | c |  } 
        Method &$k_0$ &$k_1$ & $k_2$&$k_3$& CUBS & CARS & Flowers & WikiArt & Sketch\\
         \hline
Piggyback \cite{mallya2018piggyback}&0&0&0&1&80.4&88.1&93.6&73.4&79.4\\
\hline
Piggyback$^*$ 
&0&0&0&1&80.4&87.8&93.1&72.5&78.6\\ 
Piggyback$^*$ with BN & 0&0&0&1 &82.1&90.6&95.2&74.1&79.4\\ 
Piggyback$^*$ with BN
&0&\ding{51}&0&1&81.9&89.9&94.8&73.7&79.9\\ 
        Ours (Simple, no bias)
&1&0&\ding{51}&0&80.8&90.3&96.1&73.5&80.0\\ 
        Ours (Simple) \cite{mancini2018adding}&1&\ding{51}&\ding{51}&0&82.6&91.5&96.5&74.8&80.2\\
        Ours (Simple with Sigmoid) &1&\ding{51}&\ding{51}&0&82.6&91.4&96.4&75.2&80.2\\
        Ours (Full, no bias)&1&0&\ding{51}&\ding{51}&80.7&90.2&96.0&72.0&78.8\\
        Ours (Full, no $k_2$)&1&\ding{51}&0&\ding{51}&80.6&87.5&91.0&73.0&78.4\\
        Ours (Full)&1&\ding{51}&\ding{51}&\ding{51}&82.4&91.4&96.7&75.3&80.2\\
         Ours (Full with Sigmoid)&1&\ding{51}&\ding{51}&\ding{51}&82.7&91.4&96.6&75.2& 80.2\\
        Ours (Full, channel-wise) &1&\ding{51}&\ding{51}&\ding{51}&82.0&91.0&96.3&74.8&80.0\\
        
			\hline
             
		\end{tabular}
		\label{tab:ablation-res}
        \vspace{-10pt}
\end{table*}

\begin{table*}[t]
			\caption{Impact of the parameters $k_0$, $k_1$, $k_2$ and $k_3$ of our model using the DenseNet-121 architectures in the ImageNet-to-Sketch scenario. \ding{51} denotes a learned parameter, while~$^*$ denotes \cite{mallya2018piggyback} obtained as a special case of our model.} 

		\vspace{5pt}
		\centering
		\begin{tabular}{ l | c | c | c | c | c | c | c | c | c |  } 
        Method &$k_0$ &$k_1$ & $k_2$&$k_3$& CUBS & CARS & Flowers & WikiArt & Sketch\\
         \hline
Piggyback \cite{mallya2018piggyback}&0&0&0&1&79.7&87.2&94.3&72.0&80.0\\
\hline
Piggyback$^*$ 
&0&0&0&1&80.0&86.6&94.4&71.9&78.7\\ 
Piggyback$^*$ with BN & 0&0&0&1 &81.4&90.1&95.5&73.9&79.1\\ 
Piggyback$^*$ with BN
&0&\ding{51}&0&1&81.9&90.1&95.4&72.6&79.9\\ 
        Ours (Simple, no bias)
&1&0&\ding{51}&0&80.4&91.4&96.7&75.0&79.7\\ 
        Ours (Simple) \cite{mancini2018adding}&1&\ding{51}&\ding{51}&0&81.5&91.7&96.7&75.5&79.9\\
        Ours (Simple with Sigmoid) &1&\ding{51}&\ding{51}&0&81.5&91.7&97.0&76.0&79.8\\
        Ours (Full, no bias)&1&0&\ding{51}&\ding{51}&80.2&91.1&96.5&75.1&79.2\\
        Ours (Full, no $k_2$)&1&\ding{51}&0&\ding{51}&79.8&87.2&91.8&73.2&78.1\\
        Ours (Full)&1&\ding{51}&\ding{51}&\ding{51}&81.7&91.6&96.9&75.7&79.9\\
         Ours (Full with Sigmoid)&1&\ding{51}&\ding{51}&\ding{51}&82.0&91.7&97.0&76.0&79.9\\
        Ours (Full, channel-wise) &1&\ding{51}&\ding{51}&\ding{51}&81.4&91.6&96.5&75.5&79.9\\
        
			\hline
             
		\end{tabular}
		\label{tab:ablation-dense}
        \vspace{-10pt}
\end{table*}

\subsection{Parameter Analysis}
\begin{figure*}[!b]
\centering
 \includegraphics[width=1.\textwidth,trim=2cm 0 1.5cm 0,clip]{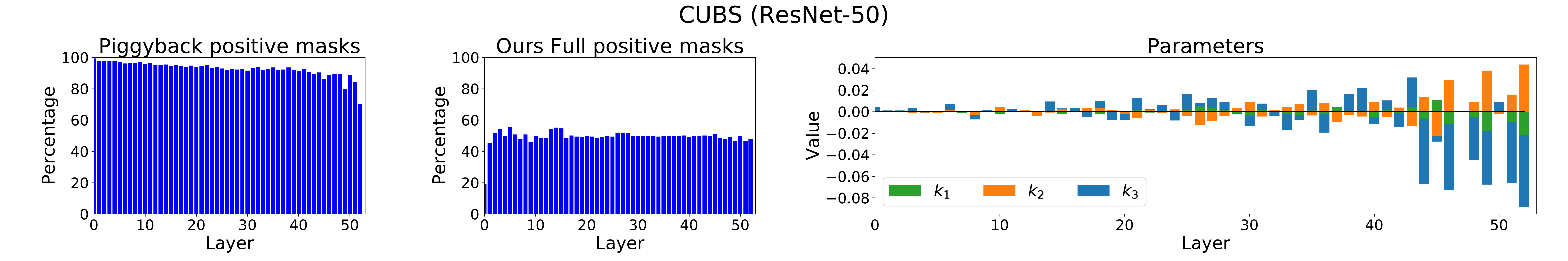}  \\
\includegraphics[width=1.\textwidth,trim=2cm 0 1.5cm 0,clip]{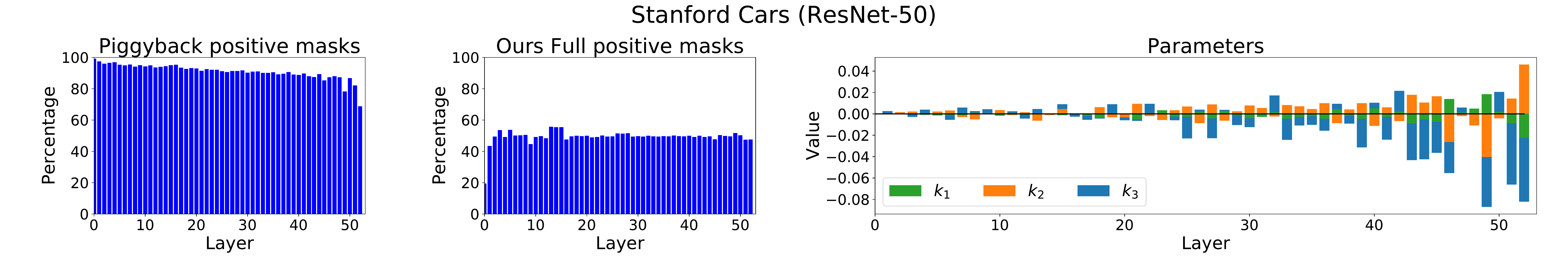} \\
\includegraphics[width=1.\textwidth,trim=2cm 0 1.5cm 0,clip]{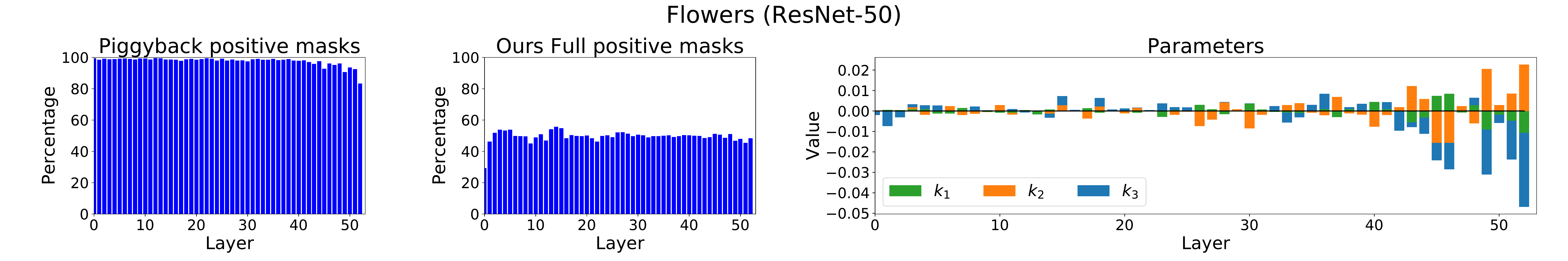} \\
\includegraphics[width=1.\textwidth,trim=2cm 0 1.5cm 0,clip]{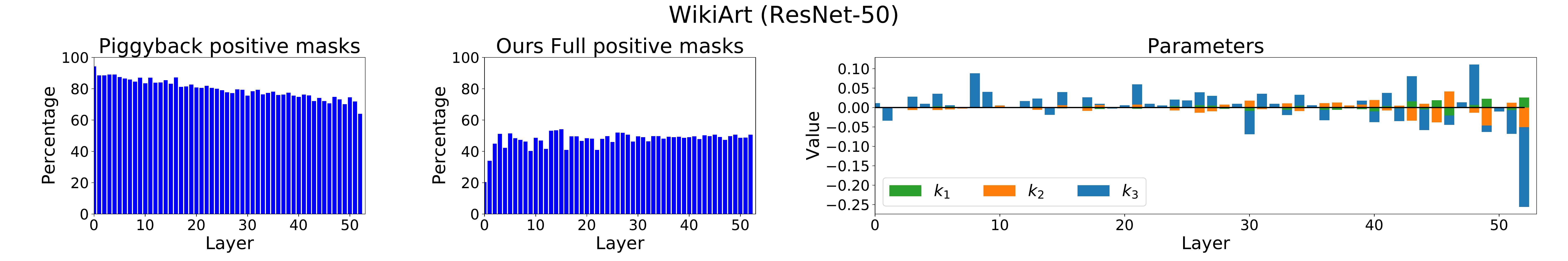} \\
\includegraphics[width=1.\textwidth,trim=2cm 0 1.5cm 0,clip]{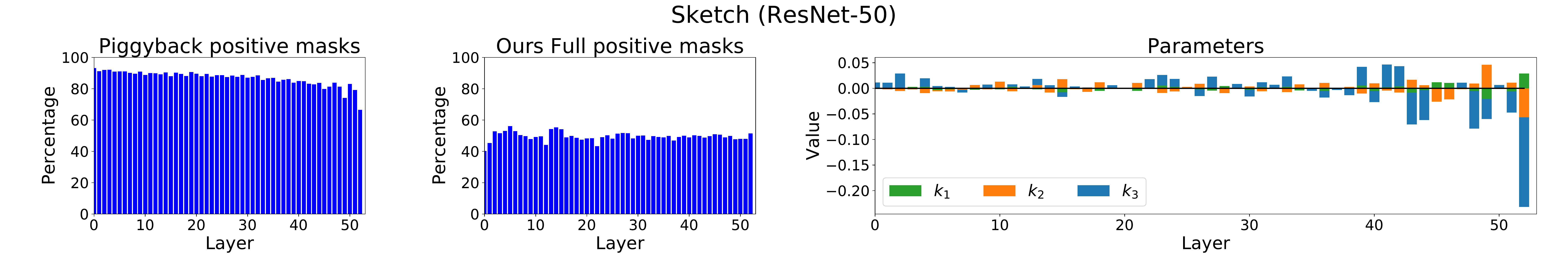} \\
  \caption{Percentage of 1s in the binary masks at different layers depth for Piggyback (left) and our full model (center) and values  of the parameters $k_1$, $k_2$, $k_3$ computed by our full model (right) for all datasets of the Imagenet-to-Sketch benchmark and the ResNet-50 architecture.
  }
  \label{fig:res-params}
  \end{figure*}

\begin{figure*}[!b]
\centering
   \includegraphics[width=1.\textwidth,trim=2cm 0 1.5cm 0,clip]{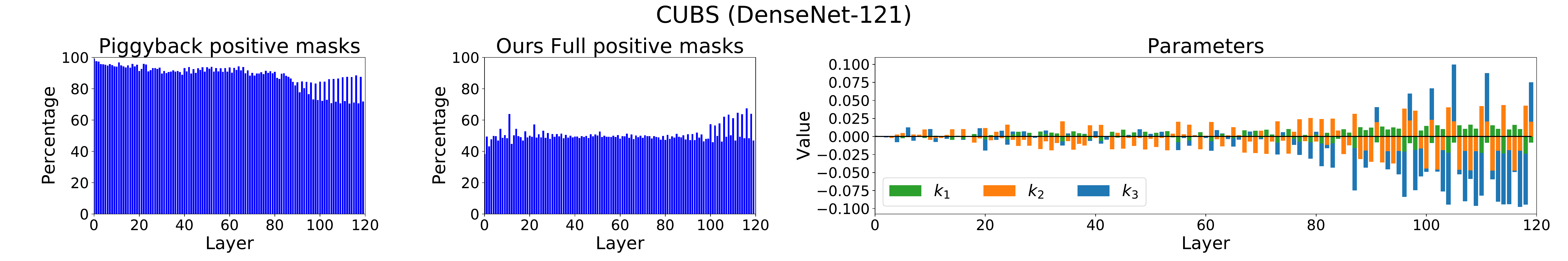}  \\
\includegraphics[width=1.\textwidth,trim=2cm 0 1.5cm 0,clip]{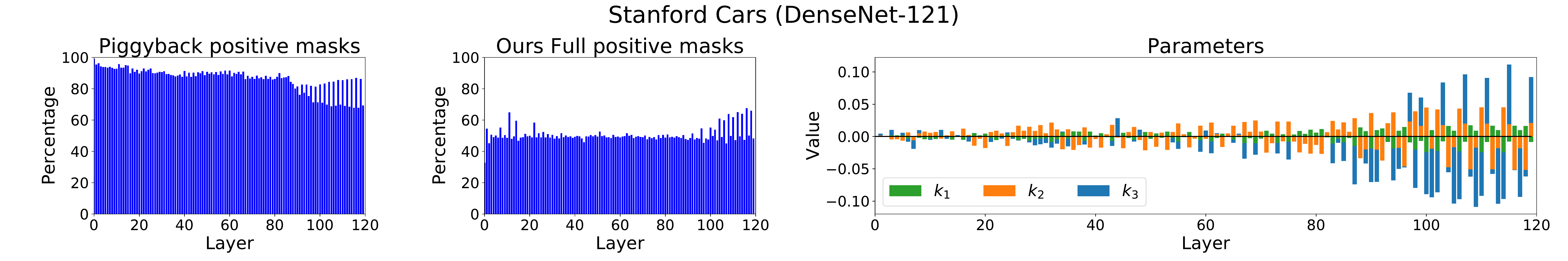} \\
\includegraphics[width=1.\textwidth,trim=2cm 0 1.5cm 0,clip]{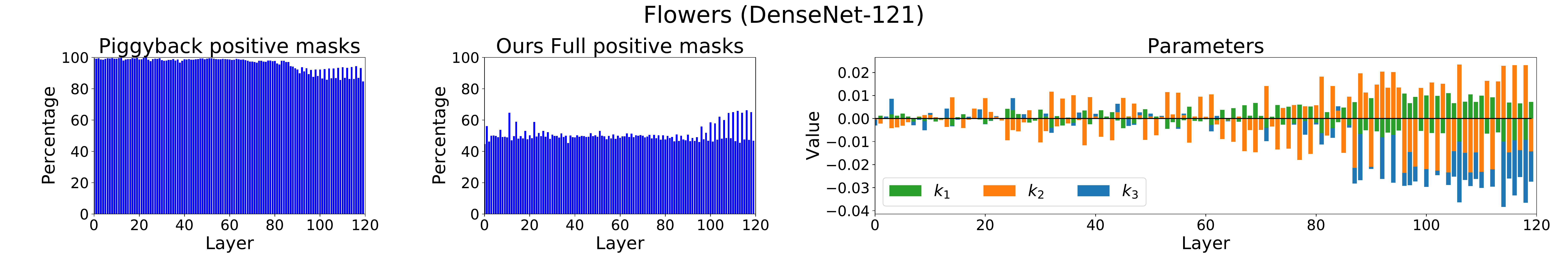} \\
\includegraphics[width=1.\textwidth,trim=2cm 0 1.5cm 0,clip]{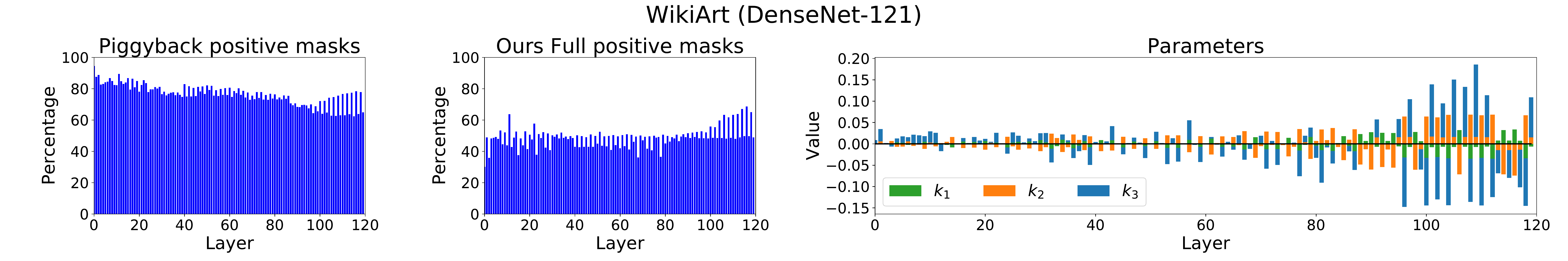} \\
\includegraphics[width=1.\textwidth,trim=2cm 0 1.5cm 0,clip]{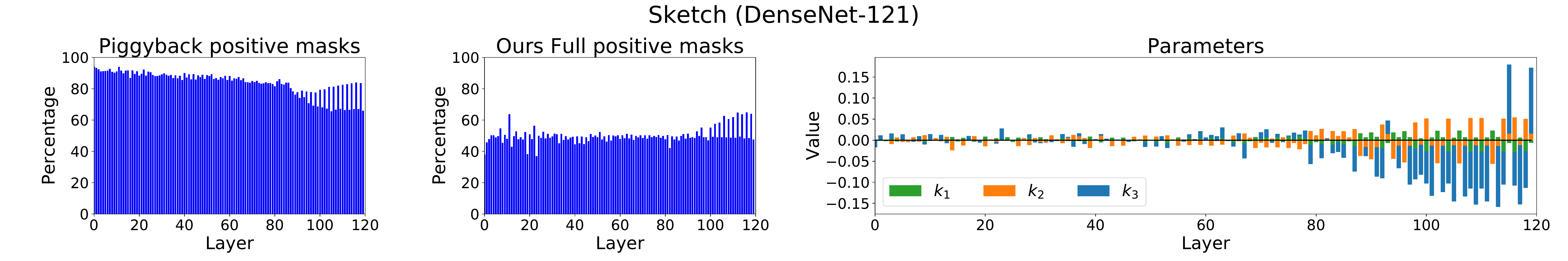} \\
  \caption{Percentage of 1s in the binary masks at different layers depth for Piggyback (left) and our full model (center) and values  of the parameters $k_1$, $k_2$, $k_3$ computed by our full model (right) for all datasets of the Imagenet-to-Sketch benchmark with the DenseNet-121 architecture.
  }
    \label{fig:dense-params}
  \end{figure*}

We analyze the values of the parameters $k_1$, $k_2$ and $k_3$ of one instance of our full model in the ImageNet-to-Sketch benchmark. We use two of the architectures employed in that scenario, \ie ResNet-50 and DenseNet-121, and we plot the values of $k_1$, $k_2$ and $k_3$ as well as the percentage of 1s present inside the binary masks for different layers of the architectures. Together with those values, we report the percentage of 1s for the masks obtained through our implementation of Piggyback. Both models have been trained considering domain-specific batch-normalization parameters. The results are shown in Figures \ref{fig:res-params} and \ref{fig:dense-params}. In all scenarios, our model keeps almost half of the masks active across the whole architecture. Compared to the masks obtained by Piggyback, there are 2 differences: 1) Piggyback exhibits denser masks (\ie with a larger portion of 1s), 2) the density of the masks in Piggyback tends to decrease as the depth of the layer increases. Both these aspects may be linked to the nature of our model: by having more flexibility through the affine transformation adopted, there is less need to keep active large part of the network, since a loss of information can be recovered through the other components of the model, as well as constraining a particular part of the architecture. For what concerns the value of the parameters $k_1$, $k_2$ and $k_3$ for both architectures $k_2$ and $k_3$ tend to have larger magnitudes with respect to $k_1$. Also, the values of $k_2$ and $k_1$ tend to have a different sign, which allows the term $k_1\mat 1+k_2\mat M$ to span over positive and negative values. We also note that the transformation of the weights is more prominent as the depth increases, which is intuitively explained by the fact that the baseline network requires stronger adaptation to represent the higher-level concepts of different domains. This is even more evident for WikiArt and Sketch due to the variability that these datasets contain with respect to standard natural images.

\section{Conclusions}\label{sec:conclusions}
This work presents a simple yet powerful method for extending a pretrained deep architecture to novel visual domains. In particular, we generalize previous works on multi-domain learning applying binary masks to the original weights of the network \cite{mallya2018piggyback,mancini2018adding} by introducing an affine transformation that acts upon such weights and the masks themselves. Our generalization allows implementing a large variety of possible transformations, better adapting to the specific characteristic of each domain. These advantages are shown experimentally on two public benchmarks, fully confirming the power of our approach which fills the gap between the binary mask based and state-of-the-art methods on the Visual Decathlon Challenge. 

Future work will explore the possibility to exploit this approach on several life-long learning scenarios, from incremental class learning \cite{RebuffiKSL17,li2017learning} to open-world recognition \cite{bendale2015towardsowr,mancini2019knowledge,fontanel2020boosting}. {Moreover, while we assume to receive the new domains one by one in a sequential fashion, our current model tackles each visual domain independently. To this extent, an interesting research direction would be exploiting the relationship between different domains through cross-domain affine transformations, to force the model to reuse previous knowledge collected from different domains.}

\begin{acknowledgements}
We acknowledge financial support from ERC grant 637076 - RoboExNovo and project \textit{DIGIMAP}, grant 860375, funded by the Austrian Research Promotion Agency (FFG).
\end{acknowledgements}


\bibliographystyle{spmpsci}
\bibliography{egbib}

\end{document}